\documentclass{article}

\usepackage{PRIMEarxiv}

\usepackage[utf8]{inputenc} 
\usepackage[T1]{fontenc}    
\usepackage{hyperref}       
\usepackage{url}            
\usepackage{booktabs}       
\usepackage{amsfonts}       
\usepackage{nicefrac}       
\usepackage{microtype}      
\usepackage{lipsum}
\usepackage{fancyhdr}       
\usepackage{graphicx}       
\graphicspath{{media/}}     
\usepackage{longtable}
\usepackage{multirow}

\newcommand{\specialcell}[2][l]{%
	\begin{tabular}[#1]{@{}l@{}}#2\end{tabular}}

\title{Build a Robust QA System with Transformer-based Mixture of Experts
}

\author{
  Yu Qing Zhou \\
  Stanford University \\
  \texttt{ivanz@stanford.edu} \\
   \And
  Xixuan Julie Liu \\
  Stanford University \\
  \texttt{xl99@stanford.edu} \\
   \And
  Yuanzhe Dong \\
  Stanford University \\
  \texttt{yzd@stanford.edu} \\
}

\begin{document}
\maketitle

\begin{abstract}
In this paper, we aim to build a robust question answering system that can adapt to out-of-domain datasets. A single network may overfit to the superficial correlation in the training distribution, but with a meaningful number of expert sub-networks, a gating network that selects a sparse combination of experts for each input, and careful balance on the importance of expert sub-networks, the Mixture-of-Experts (MoE) model allows us to train a multi-task learner that can be generalized to out-of-domain datasets. We also explore the possibility of bringing the MoE layers up to the middle of the DistilBERT \cite{sanh2020distilbert} and replacing the dense feed forward network with a sparsely-activated switch FFN layers, similar to the Switch Transformer \cite{fedus2021switch} architecture, which simplifies the MoE routing algorithm with reduced communication and computational costs. In addition to model architectures, we explore techniques of data augmentation including Easy Data Augmentation (EDA) and back translation, to create more meaningful variance among the small out-of-domain training data, therefore boosting the performance and robustness of our models. In this paper we show that our combination of best architecture and data augmentation techniques achieves a 53.477 F1 score in the out-of-domain evaluation, which is a 9.52\% performance gain over the baseline. On the final test set, we reported a higher 59.506 F1 and 41.651 EM. We successfully demonstrate the effectiveness of Mixture-of-Expert architecture in a Robust QA task. 
\end{abstract}

\keywords{Artificial Intelligence \and Natural Language Processing \and Machine Learning}

\section{Introduction}
In the task of question answering (QA), a model will be given a question as input, together with a long paragraph as context. It is expected to output an answer to the question. There are a wide variety of question types, including why, what, how, fact-based, semantic-based, etc. Specifically for our task, the model needs to select a span of text (starting and ending indexes) from the context paragraph as an answer to the question, if the question is answerable, and output N/A otherwise. 

Robustness to out-of-distribution data is critical for building generalizable NLP systems since train and test data often come from distinct user interactions or sources. In this paper, we are provided with three primary in-domain reading comprehension datasets (Natural Questions \cite{kwiatkowski-etal-2019-natural}, NewsQA \cite{trischler-etal-2017-newsqa} and SQuAD \cite{rajpurkar2016squad}) and three small out-of-domain reading comprehension datasets (RelationExtraction \cite{levy-etal-2017-zero}, DuoRC \cite{saha2018duorc}, RACE \cite{lai2017race}) for training a QA system which will be evaluated on test examples from out-of-domain datasets. 

Given the variation of QA tasks required, we aim to build a multitask language learner for our paper. And the Mixture-Of-Experts (MoE) technique, which aims to divide a complex task into appropriate subtasks, each of which can be solved by an expert network, seems to be an intuitive approach. Along with the mixture weight for these expert models produced by a gating function, we can have a QA model that can extrapolate better at each example at inference.

In this paper, we explore two architectures on the DistilBERT backbone, MoE and Switch Transformer, and train the models on in-domain datasets and augmented out-of-domain datasets to improve domain adaptive QA performances. We conduct and analyze extensive experiments to understand the effectiveness of our methods and reach the best combination of our models and techniques through our ablation study.

\section{Related Work}
While a single network may overfit to the superficial distribution in the in-domain training data, with a meaningful number of expert sub-networks, a gating network that selects a sparse combination of experts for each input example, and careful balance on the importance of expert sub-networks, a Mixture-of-Experts (MoE) model \cite{shazeer2017outrageously} can train a robust learner that can be generalized to out-of-domain datasets. However, the paper \cite{shazeer2017outrageously} does not touch on how well MoE applies to the QA task.

Inspired by the success of large-scale Transformer \cite{vaswani2017attention}, while seeking greater computational efficiency, Switch Transformer \cite{fedus2021switch} is proposed as a sparsely-activated expert model. It activates a subset of the neural network weights for each incoming example. Switch Transformer simplifies the MoE routing algorithm with reduced communication and computational costs.
    
In addition to novel architectures, data augmentation can also boost performance and robustness of training. Easy data augmentation (EDA) techniques \cite{wei2019eda}, including synonym replacement, random deletion, random swap, and random insertion, have shown effectiveness on small datasets, despite their simplicity. Back translation is another technique that has also been shown to improve reading comprehension performance \cite{yu2018qanet}, thus gaining popularity.

\section{Approach}
\begin{figure}[h!]
    \centering
    \includegraphics[width=12cm]{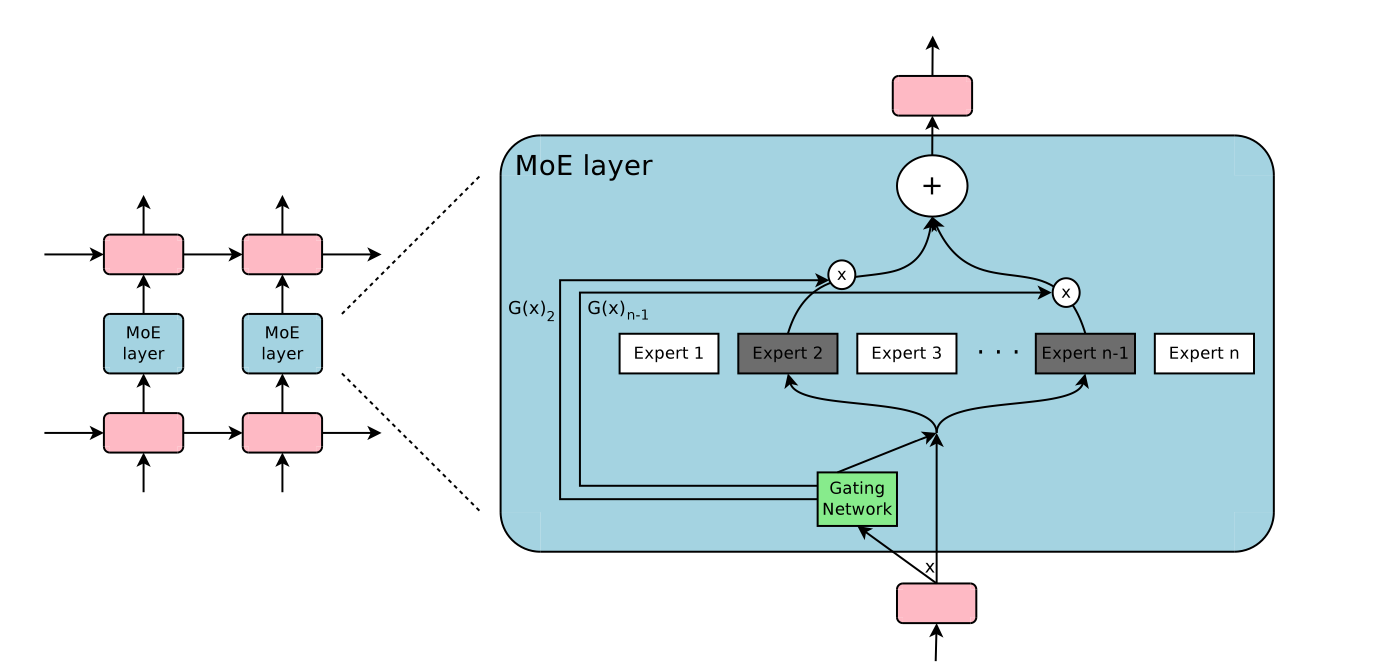}
    \caption{The Sparsely-Gated Mixture-of-Experts architecture \cite{shazeer2017outrageously} }
    \label{fig:moe_architecture}
\end{figure}

As figure \ref{fig:moe_architecture} shows, after the output layer of the DistilBERT \cite{sanh2020distilbert}, we adds $n$ single fully-connected layer in parallel as experts and another linear layer that serves as the gating function, before producing the final output. Given a input $x$, the output $y$ of the model is $y = \sum_{i=1}^{n}G(x)_i E_i(x)$, where  $G(x)_i$ is output of the gating function and and $E(x)$ is the output of $i$th expert network. 
    
\begin{figure}[h!]
    \centering
    \includegraphics[width=12cm]{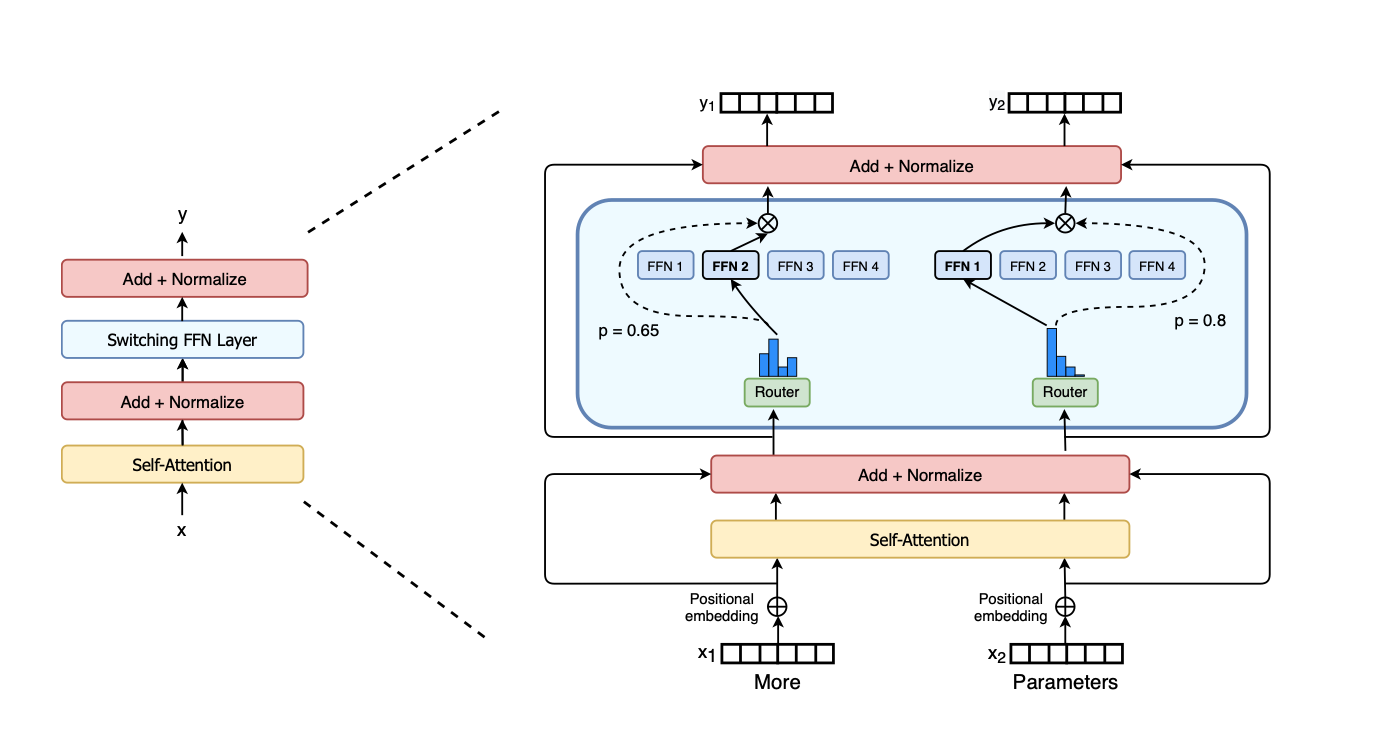}
    \caption{Switch transformer architecture \cite{fedus2021switch}}
    \label{fig:switch_transformer_architecture}
\end{figure}

For the Switch Transformer, we bring the MoE layers up to the middle of the DistilBERT model \cite{sanh2020distilbert} and replace the dense feed forward network with a sparsely-activated switch FFN layers, as figure \ref{fig:switch_transformer_architecture} shows. Through testings, we find that 8 switch FFN layers work the best. We choose 8 switch FFN layers in the following experiments.

For EDA, after data augmentation of each context paragraph, we rematch the answers within the augmented context. In order to reduce failure of rematch, we avoid operations on words within contexts that also appear in the answers.

Similarly, for back translation, we only translate context before and after answers. We use Google Translation API for its better speed and accuracy. We use Spanish, French, and German as intermediate languages (Appendix A.1). 

\section{Experiments}

\subsection{Data}
There are 6 datasets in total. Three in-domain datasets, SQuAD \cite{rajpurkar2016squad}, NewsQA \cite{trischler-etal-2017-newsqa}, and Natural Questions \cite{kwiatkowski-etal-2019-natural}, are primarily used to train a QA system; three out-of-domain datasets, DuoRC \cite{saha2018duorc}, RACE \cite{lai2017race}, and RelationExtraction \cite{levy-etal-2017-zero}, are used for evaluation. During training, the model will see 50,000 training examples from each in-domain dataset and only 127 examples from each out-of-domain datasets for fine-tuning. In the end, we will report performance on the test sets from three out-of-domain datasets.

\subsection{Evaluation method}
For evaluation, we will report the performance on both Exact Match (EM) and F1 score, averaged across the entire evaluation dataset. 
\begin{itemize}
    \item Exact Match: it is a binary measure whether the model prediction matches exactly with the target answer. 
    \item F1 score: it is calculated based on precision and recall by comparing the model prediction with the target answer word-by-word. 
\end{itemize}

\subsection{Experimental details}
From our experiments, we find that all of our models will converge within 5 epochs. Therefore, we train our models for 5 epochs, with a learning rate of 3e-5. We use a batch size of 16. 
\begin{itemize}
    \item For the Sparse-gated Mixture-of-Expert (MoE) model, we use hidden dimension of 3,072 and evaluated the number of experts in the range from 1 to 16. 
    \item For the Switch Transformer, we explore the number of transformer layers in the range from 1 to 16. 
\end{itemize}

We report the out-of-box performance of the baseline model that is trained only with the in-domain dataset. For other experiments, we use a combination of the in-domain examples and the out-of-domain ones for training. 

The data augmentation is only applied to the out-of-domain training examples, given the disproportional dominance of the in-domain examples in the training set. 

\subsection{Results}
We share a summary of our experiment results on different MoE model architecture and data augmentation techniques in Table \ref{tab:ablation}. First, with the baseline DistilBERT model, we improves the F1 by simply including the Out-of-Domain examples in the training set.

In the comparison between the MoE architecture, we find both Sparsely-gated MoE and the Switch Transformer achieves better performance over the DistilBERT baseline. The Switch Transformer pushes the performance by 3.222 to 52.052! 

\begin{table}[h!]
    \centering
    \caption{An ablation study of model architectures and data augmentation. The performance reported is the F1 achieved on the out-of-domain validation dataset. The column `Improvement' indicates the improvement over baseline)}.
    \label{tab:ablation}
    \begin{tabular}{llcc}
    \hline
    Treatment & Experiment & F1 & Improvement \\
    \hline
    \multirow{2}{*}{\specialcell{Baseline}} & DistilBERT Baseline & 48.83 & - \\
    &DistilBERT +OOD & 51.330 & 2.5 \\
    \hline
    \multirow{3}{*}{\specialcell{Explore MoE \\Architecture}} & One Expert per Dataset & 47.096 & -1.734 \\
    & Sparsely-gated MoE & 51.901 & 3.071 \\
    & Switch Transformer & \textbf{52.052} & \textbf{3.222} \\
    \hline
    \multirow{3}{*}{\specialcell{Data Augmentation\\(with Switch Transformer)}} & EDA & 52.396 & 3.566 \\
    & Back translation & 52.905 & 4.075 \\
    & EDA + back translation & \textbf{53.477} & \textbf{4.647} \\
    \hline
    \end{tabular}
\end{table}

Notably, we tried training a separate DistilBERT model for each of the datasets, and a small MLP as the gating function. This approach fails in comparison with the baseline model. To understand this, we looked into the performance of each of the separate DistilBERT model. As shown in Table \ref{tab:one_model_per_dataset}, all of the models trained on the single dataset significantly overfit to the in-domain dataset and therefore underperform in the out-of-domain examples on the validation set. The best out-of-domain F1 is only 43.469, from the model trained on NewsQA. After combining three models as experts and fine-tuning the gating function with the out-of-domain training examples, the new model gets a better F1 on the out-of-domain validation set, 47.096, but still underperform the Baseline because none of the them perform well on the out-of-domain examples. This is a mixture-of-"non-experts". In comparison, for both Sparsely-gated MoE and Switch Transformers, both of them are exposed to all 6 different datasets at training, so the inside experts are delegated to learn different underlying distributions among the dataset and the gating function is trained to select the right experts for each input example. This "self-supervised" training mechanism enables them to be generalized at diverse QA tasks, thus being more robust to domains shift. 

\begin{table}[h!]
    \centering
    \caption{The performance of the DistilBERT model trained separately on each of the in-domain dataset, reported on both the in-domain and the out-of-domain validation sets.}
    \label{tab:one_model_per_dataset}
    \begin{tabular}{llcc}
    \hline
    Training Dataset  & In-domain F1 & Out-of-Domain F1 \\
    \hline
    NewsQA & 55.66 & 43.469 \\
    SQuAD & 54.046 & 42.126 \\
    Natural Questions & 57.058 & 39.813 \\
    \hline
    \end{tabular} 
\end{table}

From the comparison of different MoE architecture, we find that the Switch Transformer gives the best out-of-domain F1. Then, we evaluate the effectiveness of data augmentation techniques with the Switch Transformers. We observe an improvement of 0.344 and 0.853 respectively with Easy Data Augmentation (EDA) and back translation. When they are applied together with the Switch Transformer, we see an even higher F1 score of 53.477. This means different data augmentation techniques can complement to each other. In future work, we recommend exploring other different data augmentation techniques to see if the performance can be lifted to a higher level. 

The combination of our best MoE architecture and data augmentation achieves a 53.477 F1 score in the out-of-domain validation set, which is a 9.52\% performance gain over the baseline. On the final test set, we reported a higher 59.322 F1 and 41.995 EM. This effectively shows the robustness of our QA system.

\subsubsection{Quantitative Analysis on the Number of Experts}\label{sec:num_experts}
Here we are showing more detailed quantitative analysis around the expert numbers in different MoE architectures. 

\begin{figure}[h!]
    \centering
    \includegraphics[width=10cm]{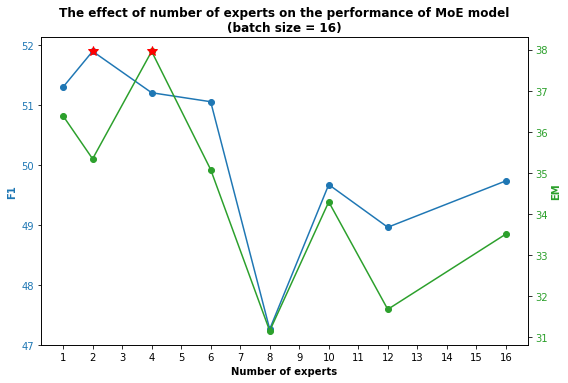}
    \label{fig:experts_in_moe}
    \caption{The performance achieved at different number of experts in the Sparse-gated Mixture-of-Expert model, evaluated in the out-of-domain F1 and EM metrics.}
\end{figure}

First, we look into the simple Sparse-gated Mixture-of-Expert architecture. In Figure \ref{fig:experts_in_moe}, we look into the effect on performance by the number of experts in the network. Compared with the minimum 1 expert, the model achieves a better F1 with 2 experts and a better EM with 4 experts. This is a good indication of adding extra experts help improves the robustness. However, the performance begins degrading with more experts added to the model. We believe it is due to the fact that the model establish over-reliance on the same few experts and the other experts only add noise instead of valuable opinions for the outputs. In the beginning of training, a few lucky experts produce good results, so they get more favored by the gating function and their parameteres are updated more often, thus reinforcing this imbalance. A possible fix to this problem is to apply more constraint to balance the importance and workloads among experts in the loss function, which we will dive deep in section \ref{sec:load_balancing}.

\begin{figure}[h!]
    \centering
    \includegraphics[width=12cm]{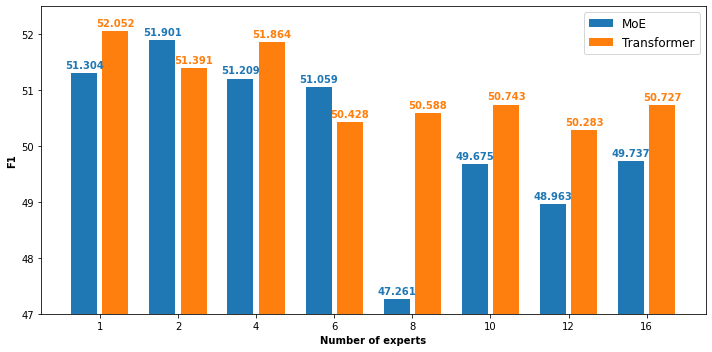}
    \caption{Performance of MoE vs Transformer, with different number of experts}
    \label{fig:moe_transformer}
\end{figure}

We also look into the effect of the expert numbers in the Switch Transformer, as shown in Figure \ref{fig:moe_transformer}. In comparison, the Switch Transformer faces a similar issue that the additional experts over 4 don't contribute to the performance, but compared with the sparsely gated MoE, the Switch Transformer is more robust to the addition of experts. In Switch Transformer \cite{fedus2021switch}, for each Switch layer, a new auxiliary loss is calculated to balance load and importance of experts. Based on our evaluation, it seems the auxiliary loss in the Switch Transformer is effective at balancing among experts.  

\subsubsection{Quantitative Analysis on Data Augmentation}
In this section, we dive deep into the effectiveness of data augmentation in the baseline DistilBERT, Sparsely-gated MoE, and the Switch Transformer respectively.

For Easy Data Augmentation (EDA), by default we generate 4 augmented context per input context. \textbf{sr} is the percentage of random synonym replacement within a sentence using a synonym dictionary WordNet, \textbf{rs} is the percentage of words randomly swapped positions within a sentence, \textbf{ri} is the percentage of inserting a synonym of a random word in a random position within a sentence, and \textbf{rd} is the percentage of words randomly deleted within a sentence. 

First, we trained the DistilBERT model only on the out-of-domain examples, augmented with EDA, back translation, and a combination of them. As shown in table \ref{tab:aug_distilbert}, all of these data augmentation brings meaningful addition of training dataset and improves the performance. The combination of both data augmentation brings the most performance gain. This is a very positive signal that the benefit of the data augmentation techniques can compliment to each other. 
\begin{table}[h!]
    \centering
    \caption{Data Augmentation - DistilBERT trained on OOD data only}
    \label{tab:aug_distilbert}
    \begin{tabular}{llcc}
    \hline
    EDA & Back translation & F1 & Improvement \\
    \hline
    None & None & 25.971 & - \\
    \hline
    sr = 0.3 & None & 26.416 & 0.445 \\
    sr = rs = ri = rd = 0.1 & None & 28.445 & \textbf{2.474} \\
    \hline
    None & Spanish & 30.17 & \textbf{4.199} \\
    None & Spanish, French & 29.741 & 3.77 \\
    None & Spanish, French, German & 29.231 & 3.26\\
    \hline
    sr = rs = ri = rd = 0.1 & Spanish & 30.638 & \textbf{4.667 (17.97\%)} \\
    \hline
    \end{tabular} 
\end{table}
    
In the experiment with the Sparsely-gated MoE, we evaluated the effectiveness of EDA. 
\begin{table}[h!]
    \centering
    \caption{Data Augmentation - EDA}
    \label{tab:aug_moe}
    \begin{tabular}{llcc}
    \hline
    Model \#experts & EDA & F1 & Improvement \\
    \hline
    MoE 1 & sr = 0.3 & 52.599 & \textbf{1.387} \\
    MoE 2 & sr = 0.3 & 51.617 & \textbf{0.101} \\
    \hline
    \end{tabular} 
\end{table}

Finally, on the Switch Transformer, we apply the back-translation through Spanish and report the performance in Table \ref{tab:aug_switch}. For almost all the expert numbers, with only one exception, the back translation improves the performance compare the counter-part without back translation. The biggest gain is observed at the Switch Transformer with 16 experts, likely because that model has the largest capacity among all. Based on these results, we add back-translation to our final model configuration. 

Interestingly, on Distilbert (table \ref{tab:aug_distilbert}), back translation with only Spanish worked better than with multiple intermediate languages (Spanish+French, or Spanish+French+German). This is likely due to over-fitting from multiple versions of back translated training data. However, in our final model configuration Switch Transformer, using three intermediate languages (Spanish, French, German) led to better performance than Spanish alone (table \ref{tab:ablation}). This shows that a model with larger capacity can benefit better from larger scale data augmentation. 

\begin{table}[h!]
    \centering
    \caption{Data Augmentation - Back translation}
    \label{tab:aug_switch}
    \begin{tabular}{llcc}
    \hline
    Model \#experts & Back translation & F1 & Improvement \\
    \hline
    transformer layers = 8\\
    \hline
    Transformer 1 & Spanish & 52.599 & \textbf{0.547} \\
    Transformer 2 & Spanish & 51.617 & \textbf{0.226} \\
    Transformer 4 & Spanish & 51.719 & -0.145 \\
    Transformer 8 & Spanish & 51.706 & \textbf{1.118} \\
    Transformer 10 & Spanish & 50.816 & \textbf{0.073} \\
    Transformer 12 & Spanish & 50.73 & \textbf{0.447} \\
    Transformer 16 & Spanish & 51.888 & \textbf{1.161} \\
    \hline
    \end{tabular}
\end{table}

\subsubsection{Quantitative Analysis on Switch Transformer's Load Balancing Loss} \label{sec:load_balancing}

Since in section \ref{sec:num_experts} we suspect that imbalance of experts' importance during training could be a reason why models with 1 or 2 expert(s) perform the best, we decided to further investigate the load balancing loss within the switch transformer, which is an auxiliary loss introduced to encourage a balanced load across experts \ref{fig:switch_transformer_architecture}. For each Switch layer, given $N$ experts indexed by $i = 1$ to $N$ and a batch $B$ with $T$ tokens, the load balancing loss is computed as the scaled dot-product between vectors $f$ and $P$:

\begin{equation}
\label{eq:load_balance}
loss_{load} = \alpha N \cdot \sum_{i=1}^{N} f_{i} \cdot  P_{i}
\end{equation}

where $f_i$ is the fraction of tokens dispatched to expert $i$ and $P_i$ is the fraction of the router probability allocated for expert $i$. The equation \ref{eq:load_balance} encourages uniform routing of the batch of tokens across the N experts. The hyperparameter $\alpha$ is a multiplicative coefficient for the auxiliary loss.

For the experiments above, we used the default $\alpha = 0.01$. Now we want to see if performance of models with 4 and 16 experts improves with larger load balancing loss coefficient $\alpha \in \{0.1, 0.05, 1, 2\}$ . Models are trained with data augmentation.

\begin{table}[h!]
    \centering
    \caption{Performance of transformer models (out-of-domain F1 evaluation scores), with different load balancing loss coefficient}
    \label{tab:balancing_loss}
    \begin{tabular}{lccccc}
    \hline
    Load balancing loss coefficient $\alpha$ & 0.05 & 0.1 & 1 & 2 \\
    \hline
    Transformer 4 & \textbf{52.828} & 52.697 & 52.351 & 49.741 \\
    Transformer 16 & 50.441 & 50.909 & 51.033 & \textbf{51.325}\\
    \hline
    \end{tabular} 
\end{table}

If we do row-wise comparison, by increasing the coefficient $\alpha$, we observe that the performance of the model with 4 experts does not improve, but the model with 16 experts does. We believe that is because with many experts, the model can spread out the load among different experts and leverage multiple experts more easily. 

However, the best performance in Table \ref{tab:balancing_loss} is achieved with 4 experts and a smaller coefficient $\alpha=0.05$. It is still smaller than the best performance in the Table \ref{tab:ablation}, which is achieved with 1 expert. The coefficient $\alpha$ provides a trade-off between ensuring load balance and the primary cross-entropy objective. It seems that this load balancing loss is effective at spreading the load among experts and involve more experts into the task, but it does not improve on the final task. A better load balancing technique would be needed for this robust Q\&A task.

\section{Analysis}

We did qualitative evaluation by reviewing our model's prediction to the out-of-domain examples and compare them with the corresponding labeled answers. Overall, we find our system provide reliable and reasonable answers to most of the context-questions pairs. It is especially good at answering fact-based questions, as shown in the examples list below. For these questions, there is a single, unique answer that can be found in the context paragraph. Our  model is able to identify the answer from the context, thus offers predictions that exactly matches with the expected answers. 

\begin{itemize}
    \item Which chromosone can you find Bcl-2?
    \begin{itemize}
        \item Prediction: kidney transplant
        \item Answer: kidney transplant
    \end{itemize}
    \item What do new users of Facebook need to create an account?
    \begin{itemize}
        \item Prediction: email address
        \item Answer: email address
    \end{itemize}
    \item What's the location of the project?
    \begin{itemize}
        \item Prediction: Arizona desert
        \item Answer: Arizona desert
    \end{itemize}
    \item What is the name of Boris Diaw's team?
    \begin{itemize}
        \item Prediction: San Antonio Spurs
        \item Answer: San Antonio Spurs
    \end{itemize}
    \item Where did Hasumi receive his MBA?
    \begin{itemize}
        \item Prediction: Harvard University
        \item Answer: Harvard University
    \end{itemize}
\end{itemize}

There are also several examples that we found in which either multiple answers should be allowed, but because only a single answer is provided in the label, our model is falsely penalized. For example, for the first question below, we believe that the third stage and the withdrawal stage are equivalent, based on the context, but because only "withdraw stage" is provided in the label, the prediction is considered as 0 EM and 0.5 F1. Similarly, in the second example, both "gray haze" and "smog" are the same thing, but the model prediction is considered as 0 EM and 0 F1. This is the limitation of our evaluation dataset and metrics.

\begin{itemize}
    \item In which stage will people feel most uncomfortable?
    \begin{itemize}
        \item Prediction: third stage
        \item Answer: withdrawal stage
    \end{itemize}
    \item According to the news report, what does Beijing have in common with Los Angeles?
    \begin{itemize}
        \item Prediction: gray haze
        \item Answer: smog
    \end{itemize}
\end{itemize}

We also find that our model is not good at summarizing a paragraph or answering complex questions with multiple conditions. In the first example below, the question starts with "what do we know about X" -- this is about summarizing the takeaway from the context paragraph. Our model gives a half sentence that is probably extending "California sea lions" and doesn't make sense itself. Similarly, the second question is about "what does A do to B", and our model's prediction is very far from the expected answer. This indicate that our model is not trained sufficiently to answer this type of questions. We look through the in-domain training examples and find very few questions that look like this. This explains why our model is bad at this type of questions. Another type of the question that the model fails has multiple conditions, like the third example below. The question asks for the place where two conditions need to meet. Our model fails to provide the correct answer. This type of the questions require reasoning based on a long paragraph. This is a difficult task and our model does not learn well to answer this type of questions yet. 

\begin{itemize}
    \item What do we know about California sea lions?
    \begin{itemize}
        \item Prediction: are the fastest of all the
        \item Answer: Males are much larger than females
    \end{itemize}
    \item What does Dong-Jin do to Ryu ?
    \begin{itemize}
        \item Prediction: ambushes and murders the organ dealers.
        \item Answer: set up an electric booby trap on his doorknob, which renders Ryu unconscious
    \end{itemize}
    \item Where can you enjoy both convenient transport and beautiful beaches?
    \begin{itemize}
        \item Prediction: Thailand has a lot to offer, from the party-central Bangkok
        \item Answer: Melbourne, Australia
    \end{itemize}
    \item What year did Santer-Poos Ministry II start?
    \begin{itemize}
        \item Prediction: 1989 and 13 July 1994
        \item Answer: 1989
    \end{itemize}
    \item Why do Ryu and Dong Jin wait at each others' residence ?
    \begin{itemize}
        \item Prediction: attempt to kill
        \item Answer: Ryu arrives at Dong-jin's residence in an attempt to kill him
    \end{itemize}
\end{itemize}

Lastly, we find the following example very interesting. The question contains an error: it should be brother name instead of brothers name. Our model is likely confused about this error, so it returns two similar names, Constantine II and Constantius II, where as the expected answer is "Constantius II". It indicates that our model pays great attention to the words in the question and can be confused if the question contains typo or grammar mistakes.  

\begin{itemize}
    \item What is Constans's brothers name?
    \begin{itemize}
        \item Prediction: Constantine II and Constantius II
        \item Answer: Constantius II
    \end{itemize}
\end{itemize}

\section{Conclusion}
To conclude, our combination of best MoE architecture and data augmentation achieves a 53.477 F1 score, which is a 9.52\% performance gain. On the final test set, we reported a higher 59.506 F1 and 41.651 EM. We successfully demonstrate the effectiveness of Mixture-o-Expert architecture in a Robust QA task. Based on the qualitatively analysis, we find our model is very reliable and accurate at answering fact-based questions whose answers can be found from the context paragraph; it fails at questions that require reasoning or summarizing the long paragraphs. 

One limitation of our work is that we did not get the time to investigate the imbalance of experts' importance in our models. Avenues for future work could include analysis of the acquired expertise of each expert, further adjustment of loss functions in order to train experts better, and different routing mechanisms of the gating function (for example, dataset classification could be a straightforward way to direct input data to experts specialized on each dataset).

\bibliographystyle{unsrt}  
\bibliography{references}  

\appendix

\section{Appendix}

\subsection{Exploration of back translation}

Using Google API, we realized that languages closer to English as intermediate languages lead to more stability in translation.

For example, the string from training data "EETdE BTdB \$28,530,608 EETdE BTdB Memphis Grizzlies EETdE EETrE BTrB BTdB James Harden EETdE BTdB \$28,299,399 EETdE BTdB Houston Rockets EETdE EETrE BTrB BTdB DeMar DeRozan EETdE BTdB \$27,739,975 EETdE BTdB Toronto Raptors EETdE EETrE EETableE" failed to be translated through Chinese due to its unconventional words that cannot be matched in Chinese. 

When broken down to smaller phrases such as "EETdE BTdB \$28,530,608 EETdE BTdB " it was somehow able to be translated through Chinese.

The original string, however, could be translated through languages closer to English, such as Spanish or French.
\end{document}